\documentclass[letterpaper, 10 pt, conference]{IEEEtran}  
\usepackage{graphicx}
\usepackage{amsmath}
\IEEEoverridecommandlockouts

\IEEEoverridecommandlockouts                              

\title{\LARGE \bf
The ATTUNE model for Artificial Trust Towards Human Operators
}

\author{Giannis Petousakis$^{1}$, Angelo Cangelosi $^{1}$, Rustam Stolkin $^{2}$, and Manolis Chiou$^{3}$
\thanks{© 2024 IEEE.  Personal use of this material is permitted.  Permission from IEEE must be obtained for all other uses, in any current or future media, including reprinting/republishing this material for advertising or promotional purposes, creating new collective works, for resale or redistribution to servers or lists, or reuse of any copyrighted component of this work in other works}%
\thanks{*This work was undertaken in the Cognitive Robotics Lab, University of Manchester, Department of Computer Science. It is published in IEEE SMC 24.}
\thanks{$^{1}$Cognitive Robotics Lab, University of Manchester, Department of Computer Science, United Kingdom
    {\tt\small ioannis.petousakis@postgrad.manchester.ac.uk}
    {\tt\small angelo.cangelosi@manchester.ac.uk}}%
\thanks{$^{2}$Extreme Robotics Lab, University of Birmingham, School of Metallurgy and Materials, United Kingdom
    {\tt\small r.stolkin@bham.ac.uk}}%
\thanks{$^{3}$Manolis Chiou, Queen Mary University of London, United Kingdom 
    {\tt\small m.chiou@qmul.ac.uk}}%
}

\begin{document}

\maketitle
\thispagestyle{empty}
\pagestyle{empty}

\begin{abstract}

This paper presents a novel method to quantify Trust in HRI. It proposes an HRI framework for estimating the Robot Trust towards the Human in the context of a narrow and specified task. The framework produces a real-time estimation of an AI agent’s Artificial Trust towards a Human partner interacting with a mobile teleoperation robot. The approach for the framework is based on principles drawn from Theory of Mind, including information about the human state, action, and intent. The framework creates the ATTUNE model for Artificial Trust Towards Human Operators. The model uses metrics on the operator's state of attention, navigational intent, actions, and performance to quantify the Trust towards them. The model is tested on a pre-existing dataset that includes recordings (ROSbags) of a human trial in a simulated disaster response scenario. The performance of ATTUNE is evaluated through a qualitative and quantitative analysis. The results of the analyses provide insight into the next stages of the research and help refine the proposed approach.   

\end{abstract}


\section{INTRODUCTION}

With the advancements in robotics, both in terms of AI and hardware, we are seeing a rapid propagation of robotics in every aspect of our lives. This development makes the study of Human-robot interaction (HRI) and Human-Robot Teaming (HRT) increasingly necessary and relevant. 

HRI has studied the various aspects of the interaction between robots and humans, from the design and development of more engaging robots to the creation of more capable robots. In HRT  the role of the human is often that of the operator and decision maker. Through the robot, the operator can perform tasks that would otherwise be dangerous or not cost-effective.

Eventually, the capabilities of the robots and their AI improved to an extent that enabled their utilisation in a more substantial and impactful way. This led to the development of Human-Robot Collaboration (HRC) and Teaming, which allowed the robots to cooperate with humans as members of a team toward a common goal. Much research has been done to improve the AI agent's capabilities using more nuanced methods. Through these methods, the ways the AI agent can interact with their human teammates have been expanded. This gave grounds for AI agents that can better assist and interact with humans.

An intrinsic part of human teams is the fact that humans can naturally understand their partner's intent, state, and actions, helping them trust one another. This trust allows humans to collaborate more efficiently and in a complementary manner, being able to assist when needed. The next step towards that direction has been the effort to improve the capabilities of the AI agent by expanding the ways it can perceive its human partner, itself and the environment.

The end goal of the effort is the improvement of Human-Robot Interaction and Teaming. One of the big advantages of humans is the ability to collaborate with their partners as members of a team with a shared goal. This is possible because humans can interpret their partner's actions and understand their intent. This is an ability that is developed naturally \cite{c1} and has been identified as the ability to hold a Theory of Mind (ToM). Thanks to the ability for a ToM humans can trust their partners through observation of their behaviour combined with prior knowledge of their partners and the world. 

Following the same principle, in this study, we propose a framework that allows the AI agent to estimate a level of Trust towards the human. This paper does not aim to provide an approach for generalized robotic trust towards human agents. Instead, it focuses on specific and defined tasks with discrete goals and risks. The framework was applied to the task of robot teleoperation to study its effectiveness. The scenario had the operator explore and inspect a simulated disaster site. The level of AI agent Trust towards the operator was quantified by combining the data about them. 

Using the framework, the Artifical Trust Towards hUmaN opErator (ATTUNE) model is created. ATTUNE uses metrics of the human's state, intent, and actions to extract context about them in real-time. It then processes and blends this information to create the Artificial Trust metric. Using principles rooted in ToM, we aim to create and extract context from new methods, and formalize the approach and use of previously created methods. The model is tested on a pre-existing dataset that includes recordings (ROSbags) of a human trial in a simulated disaster response scenario. The performance of ATTUNE is validated through a qualitative and quantitative analysis.

The primary research goal is to conceptualize and create a framework able to produce a quantification of the level of trust towards a human operator in a specific context. In this paper we aim to answer the following questions: 

\begin{itemize}
    \item Q1 What metrics could be used for the creation of the trust metric
    \item Q2 How can we combine the various metrics to quantify a trust metric
    \item Q3 How can we evaluate the performance of the created trust estimation model 
\end{itemize}

\section{Background and Related Work}

\subsection{Trust}

While trust has received many definitions, the one that is more relevant to our case is the following: 'Trust is the Attitude that an agent (automation or another person) will help achieve an individual’s goals in a situation characterized by uncertainty and vulnerability' \cite{c2},\cite{c3}.

For our case of HRT we consider the AI agent trust or Artificial trust, the attitude that an agent will contribute to the goals of the human-robot team. 

In HRI, trust was divided into two different categories: performance-based trust and relation-based trust \cite{c4}, with performance-based trust playing the major role in the creation of trust \cite{c5},\cite{c6}. 

The primary factors affecting trust in HRI have been defined as (also see Fig. 1)\cite{c7}: 

\begin{itemize}
  \item Human Related factors
    \begin{itemize}
        \item Traits
        \item States
    \end{itemize}
  \item Robot Related factors
    \begin{itemize}
        \item Features
        \item Capability
    \end{itemize}
  \item Environment Related factors
    \begin{itemize}
        \item Team
        \item Task/Context
    \end{itemize}
\end{itemize}

Another aspect of Trust that is necessary for our approach is that of Trustworthiness, which has been defined by Gambeta \cite{c8} as follows: 'When we say we trust someone or that someone is trustworthy, we implicitly mean that the probability that he will perform an action that is beneficial or at least not detrimental to us is high enough for us to consider engaging in some form of cooperation with him.'. 

We can interpret this definition of trustworthiness as the probability that someone will act beneficially or at least not detrimentally to the goal of the human-robot team. This approach to trustworthiness is crucial for our implementation as it is a proponent of trust. Trustworthiness has been defined as a property of the Trustee and affects the Trustor's choice of who to Trust. Likewise, Trust is a property of the Trustor about the Trustee  \cite{c9}. 

The last factor affecting Trust, relevant to our method, is that of trust and reputation. While trust is usually created gradually through the observation of an agent, in the cases we lack prior information about them we can resolve to reputation. Reputation is an opinion of others regarding an individual that we can use to assess their trustworthiness if we do not have personal experience of it\cite{c10}. 

Based on these principles our approach is that the AI agent's Trust towards the human can be measured through the human’s actions, intent, state and performance while completing a certain task.

\subsection{Theory of Mind}

The next step in our process was related to processing information about the human's state, intent, and actions to introduce a measure of Trustworthiness into our model. Our approach towards that end was the use of principles and concepts rooted in Theory of Mind. 

Theory of Mind was defined as the ability of an individual to attribute mental states to themselves and others. Using this ability, an individual can make predictions about the behaviour of others \cite{c11}. Theory of Mind was studied primarily during the developmental stage of children or children on the autism spectrum which demonstrated a deficit in their ability to form a Theory of Mind, independent of general intellect levels \cite{c1}.

A Theory of Mind approach was proposed by Scasselati \cite{c12} for robotics. It was proposed that implementing concepts of Theory of Mind in the creation of robots would allow for social interactions between robots and humans. Robots would be capable of naturally learning from others, and expressing their internal states. Robots would also recognize the goals and desires of others, react to others and adapt their behaviour based on their perception of others.

\subsection{ToM and intent recognition}

An implementation of ToM relevant to our framework is the use of ToM to help with the navigational intent recognition of observed humans \cite{c13}, \cite{c14}. Through the improved understanding of the human's state, actions and intent, the AI agents can infer their intent and possibly assist in the human's goals. Towards this end, Vinanzi et al \cite{c15} created a method of reading the intention of a partner based on their body posture. 

Intention recognition has also been applied to the human's actions. In \cite{c16} the robot can decompose complex human actions into their more basic components and use those to identify and differentiate between the various goals the human might want to achieve. This implementation can allow the robot to autonomously cooperate with and assist in the human's routines.   

\subsection{ToM, and Trust in HRI}

Theory of Mind has also been used as a component for the development of trust in Human-Robot Interaction. Devin used Theory of Mind to improve the shared Human-Robot team’s plans execution \cite{c17}. The robot used a ToM Manager that allows it to estimate and maintain the mental states of each agent it’s interacting with, breaking them down into more basic components that included world states, goals, plans, and actions. 

A significant research effort has been put into identifying the characteristics that influence the trustworthiness of robots  in \cite{c25}. They studied the effect ToM had on the perception of the robot's trustworthiness and the overall impact on the HRI.

Human Trust towards robots is a major focus of work on Trust in HRI. In \cite{c18} they developed a method to predict human trust in a robotic swarm control scenario. The method allowed them to predict the trust level without asking for trust feedback values from the operator to reduce the cognitive load on them. Similarly, in \cite{c20}, they studied the effects of human trust towards a robot to evaluate the effectiveness of a robot in patient care assistance. Human trust towards the robot  \cite{c21}  was also integrated into the robot's decision-making to improve the performance of the Human-Robot Team  \cite{c22}. Conlon et al. \cite{c23} proposed a self-assessment method that provided a metric of the robot’s confidence in completing the task. This metric was presented to the Human partner to improve task performance and the Human trust towards the robot.

\subsection{Triadic Human-AI Agent-Operator Interaction}

With the advancements in robotics and AI we have started to see the appearance of a triadic relationship in HRI, with the third party introduced being the Deployer of the robot \cite{c24}. In this context, the Deployer is considered the party responsible for deploying the robot, from designing and creating the hardware to implementing its various subsystem functionality and creating its AI agent. In our case, since we are focused on the AI agent of the robot, the interaction between Deployer and the Robot can include any modules that might affect the robot's behaviour, policies, and goals. 

Our approach to creating the Artificial Trust model was done by analyzing the steps through which we could decide to trust the human operator, in the context of a specific task, through the use of the information available about them. The information available to us is directly connected to the information available to the robot and AI agent, and to primarily affected by context we can extract from this information. Using this process, we aim to create models capable of replicating our own process towards trusting a human and relegating the task to the AI agent, which will, in turn, estimate a level of Artificial trust towards the human. 

We chose to combine two approaches to build our trust model, one was purely performance-centric, which could easily be applied and interpreted by the robot. The second entailed using principles from ToM on the side of the Deployer to provide the AI agent with nuance about the human. 

\subsection{Related Work}

Currently, there is a very low volume of research in HRI that focuses explicitly on the estimation and quantification of Trust from the Robot towards the Human.  

Vinanzi et al. \cite{c26} reproduced the results of \cite{c27} using a physical robot. They compared the ability of AI agents with Mature and Immature ToM to identify tricker and helper agents. They also introduced an episodic memory module that created different characters for the robot based on its past experiences. 

In \cite{c28} they implemented a trust estimation system based on speech to evaluate the robot's trust in a human's navigational directions. While in \cite{c29} they used a partially observable Markov decision process (POMDP) to estimate trust in a human in the context of an object handover task.

Two more pieces of research that are closely related to our work are the following. In \cite{c30}, using an approach similar to our method, they created a system that can estimate a metric of robot health to identify when a robot needs assistance from the human operator. Lastly, in \cite{c31}, they used a metric on the operator's navigational intent and attention to create the Hierarchical Expert-guided Mixed-Initiative Control Switcher (HierEMICS controller). The controller was implemented to improve the Level of Autonomy switching in a Mixed-initiative control scenario to reduce the occurrence of conflict for control.

\section{Artificial Trust Framework and the ATTUNE model}

\subsection{Artificial Trust Framework}

The aim of this work is to introduce Trust in HRI from the perspective of the AI agent towards its human partners. This was done by adapting previous approaches, introducing principles based on Theory of Mind, and introducing metrics that can provide the AI agent with insight on both the human and the robot state. 

With Schaefer’s \cite{c7} approach towards trust in HRI as a starting point, we will provide the robot with information about humans, robots, and the environment. Since our approach focuses on the Trust of the AI towards the human, we expand the Human Related factors to include information about the operator's:

\begin{itemize}
  \item State
  \item Intent
  \item Actions
  \item Performance
\end{itemize}

Performance-based metrics are usually straightforward to implement. When the metrics contain information about the operator's state, intent and actions, they tend to be harder to interpret. By approaching these metrics through the scope of ToM, we can provide the AI agent with more context about the human. 

Using this approach we propose a framework for creating an AI agent that can quantify and measure its Trust towards its human partner. Trust is a product of the AI's confidence in the human's ability to contribute to the Human-Robot Team’s intended task. It is estimated in real-time through the confidence metrics on the operator’s state, intent and performance. 
These metrics represent the human's trustworthiness and are implemented through probability distribution functions. These discrete confidence metrics are then combined using a weighted average method to create the overall level AI agent Artificial Trust towards the operator. 
From this point on, in the context of this paper, we will refer to Artificial Trust as Trust, as we are only studying the one-directional trust of the AI agent towards the human. 
The trust estimation process can be adjusted by altering the probability distribution functions, and optimizing the weights used. It could also be improved by introducing new metrics to the model. 

The level of Trust is adjusted, by a coefficient dependent on any positive or negative incidents that occur due to the actions of the operator. The trust coefficient is estimated in real-time, when and if it occurs. This factor can be adjusted by changing the penalty/reward for each different incident and by introducing more incidents that contribute towards its calculation. 

Finally, an episodic memory module was used to create an episodic memory profile for short and long-term memory of the operator's Trust level. The short-term memory module stores and updates the level of trust estimated in real time. The long-term memory module is not used currently but is implemented as a representation of the operator's reputation.


\subsection{ATTUNE Model}

The proposed Artificial Trust framework was used to create the ATTUNE model (see fig. 1). The specific metrics to create each confidence estimation were based on a previous experiment \cite{c31} that had produced a detailed dataset in the form of ROSbag recordings for a large number of participants. In the context of that experiment, the Human-Robot Team was tasked with navigating through and inspecting points of interest at a high-fidelity simulated disaster site. 

\begin{figure}
	\centering
	\includegraphics[width=0.99\columnwidth]{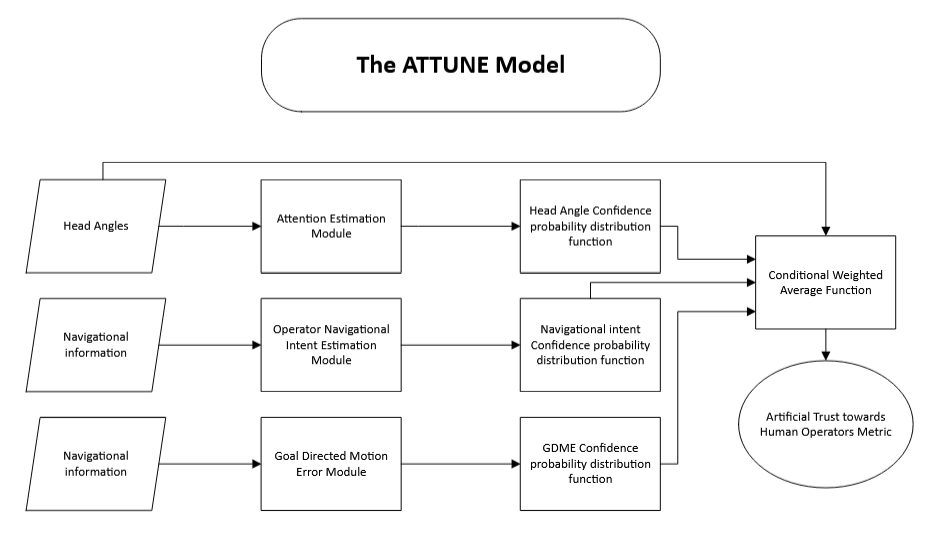}
	\caption{Diagram of the ATTUNE Model}
	\label{fig:system}
    \medskip
\end{figure}

The data includes not only information about the HR Team's performance and the robot but also information about its human partner's state of attention and navigational intent.  

The AI agent monitored human attention to the graphical user interface unit through their head angles (mainly Yaw) and their navigational intent through a Bayesian probabilistic intent estimation method.

\subsubsection{Artificial Trust Metric}

The Artificial Trust Metric combines the three confidence metrics through a conditional weighted average function. In the case of our application, the condition affected the distribution of the weights and was tied to the operator's state of attention, as inferred based on their head angles (HAngle, in the context of the conditional weighted average function). This was done to further penalize unsafe driving of the robot, i.e. when the operator was driving while not attending the GUI screen. The weights were chosen based on expert knowledge of the paradigm. We tested various sets of weights and chose weights that produced an evenly spread estimation of trust. The further optimization of the weights was beyond the scope of this implementation. 
 
\begin{equation*}
  HAngle =
    \begin{cases}
      \text{\textgreater 17 w1= 0.5 , w2=0.15 , w3 0.35}\\
      \text{else w1= 0.3 , w2=0.15 , w3 0.55}\\
    \end{cases}       
\end{equation*}

\begin{multline*}
Trust(n) = \sum_{n} (w1*confidence_h(n)\\+w2*confidence_e(n)+w3*confidence_i(n))
\end{multline*}

\subsubsection{Attention-based Confidence}

The metric of confidence based on the human State is created by monitoring the operator's state of attention. This is done through a model \cite{c32} designed to identify whether the operator is attending the user interface screen based on their head angles (ha, in the context of the estimation of the head angle confidence). In the original experiment, 17 degrees had been identified as the range of head angles above which the operator could not comfortably attend the GUI of their setup. Since the experimental setup in our dataset was the same we used the established head angle range to create the sigmoid probability distribution function. By applying this function to the data collected by the robot we estimated the confidence metric based on attention. This metric of confidence is a representation of the operator's trustworthiness based on their attention state.         

\[ confidence_h(s) = \frac{1}{(1+exp((2.5*ha-2.5*17)))} \]

\subsubsection{Navigational Intent-based Confidence}

To monitor the human-related factor of Intent we used the model created in \cite{c33}. This model can provide an estimation of the operator's navigational intent (i). By applying a sigmoid probability distribution function to the navigational intent we created the navigational-intent based confidence metric for the operator. This metric describes the confidence of the AI agent that the operator will positively contribute towards the shared goal, measuring the trustworthiness of the operator.

\[ confidence_i(i) = \frac{1}{(1+exp((-13*i+13*0.5)))} \]

\subsubsection{Performance-based Confidence}

The performance factor was monitored in real-time by the Goal-directed Motion Error metric (p) \cite{c34}. This metric was also used in \cite{c31} and produces an operator navigational error metric. This metric is calculated by comparing the current path, to the ideal robot path for ideal performance based on the knowledge of the current task. This data is applied to a sigmoid probability distribution function to create the related confidence metric, being the last trustworthiness metric of the operator. 

\[ confidence_e(p) = \frac{1}{(1+exp((11*p-11*0.5)))} \]

\subsubsection{Trust Coefficient Factor and Memory module}

Once the Artificial Trust metric was created, it was adjusted with information about the operator's actions. We introduced the trust coefficient factor (tcf) to account for any task-critical incidents caused by the operator's actions. 
This factor would be updated only when such an incident occurred and was used to augment the overall AI agent's trust. The penalties and rewards can be adjusted for each action based on its impact or risk in the context of the task. 
A positive incident would grant a reward, e.g. a successful exploration of a Goal, or Point of Interest, would produce a reward of 0.033. A negative incident would result in a penalty, e.g. a collision would reduce the trust coefficient factor by 0.2. We considered and tested various approaches for the method and rate of adjustment of the tcf, choosing to go with increments that fit the criticality and context of the given task. 

In our implementation, the artificial Trust metric was utilized through a short-term memory module where the penalty and reward were an additive adjustment to the overall artificial trust value. More explicitly, the short-term trust coefficient factor (tcf) was initialized at (0) and had a range of [-1,1].  This was done to promote faster adjustment of Trust for any incidents occurring for the duration of that run. 

\[ Short term Trust = Trust + stcf \]

On the other hand, the long-term memory model, representing the reputation of the operator, applied the penalty and reward multiplicatively to the overall trust value. The long-term trust coefficient factor (ltcf) was initialized at (1) and had a range of [0,2] This was done to ensure a more measured/ gradual adjustment of the Artificial Trust between runs.

\section{Model Evaluation}


We used data from the HierEMICS\cite{c31} experiment to do a pilot evaluation of the performance of the ATTUNE mode. The experiment had the operators navigate and inspect a disaster site simulated in high fidelity in Gazebo/ROS. The ROSbags from the human trial were replayed in real time. The ATTUNE model, subscribing to the appropriate topics created a real-time estimation for the level of AI agent Trust towards the operator in each trial.

We evaluated the ATTUNE model by comparing the following:

\begin{itemize}
    \item The ranking of the operators based on real time AI Trust estimation.
    \item The post-hoc analysis of their performance.
    \item The qualitative classification of the operator's driving behaviour.
\end{itemize} 

The post-hoc analysis of the operator's performance and incident data sorted the operators based on their capability. The capability of each operator was estimated based on their performance in three of the metrics that were available to us from the dataset. The metrics used were the total time to complete each trial, the number of collisions and the number of goals inspected successfully. We chose those three metrics to evaluate the operators as we did not want to make big assumptions between the other metrics and the estimated capability of the operator. The assumptions made are the following. A capable operator should on average be faster, show no collisions and inspect all goals.

The qualitative classification was created on the basis of the actions of the operator. The notes of the experiment were analysed and the recording of each operator's attempt was evaluated in a fashion similar to a supervisor monitoring them. This analysis produced an approximate stratification or grouping of the operators. The operators were assigned, based on their level of competence, to the following categories: above-average, average, and below-average. 

\section{Results}

The levels of trust produced by the ATTUNE model agree, for the most part, with the data-driven and qualitative analysis of the ranking of the operators, ordering them based on their performance.

The operators that the ATTUNE model placed in the lower estimated trust levels were the only ones that had collisions and missed goals. They also showed incidents of dangerous or hectic driving. On average, these operators also showed longer times to complete the task and an increased amount of backtracking compared to the operators that were considered more trustworthy. On the other hand, the operators placed in the higher strata had no missed goals and no collisions, and demonstrated more confident driving and navigation of the arena.

\begin{figure}
	\centering
	\includegraphics[width=0.99\columnwidth]{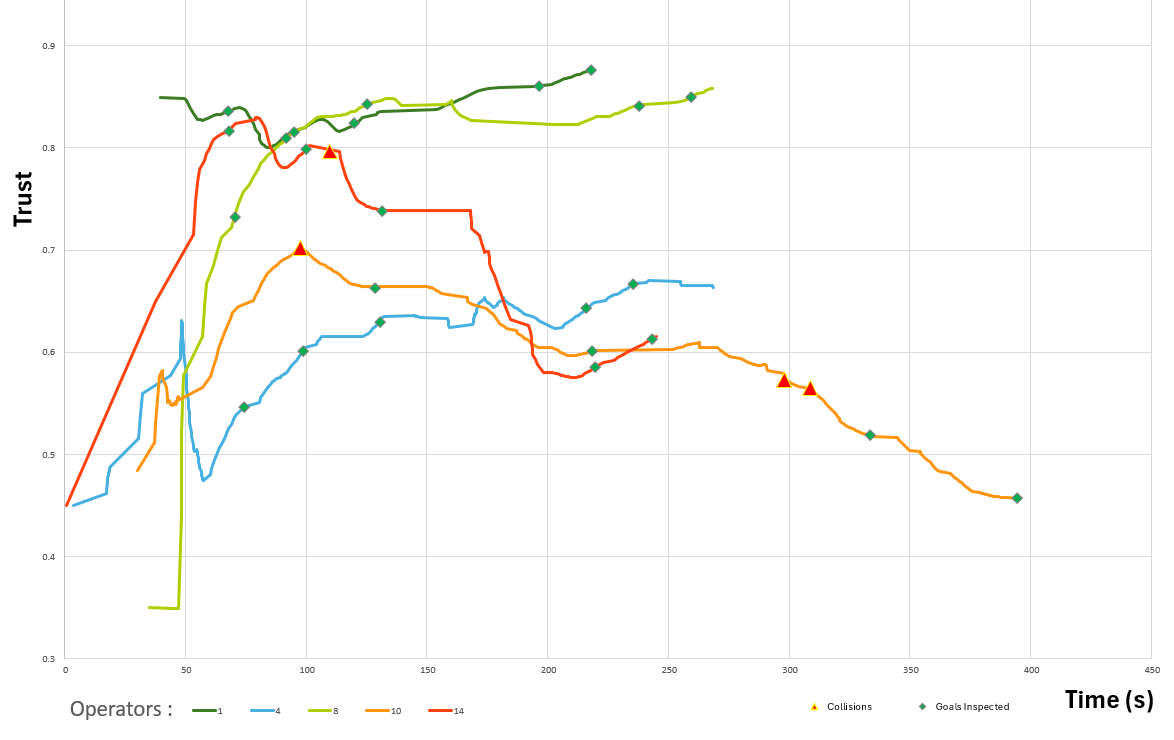}
	\caption{Ranking of the Operators based on the AI agent Trust}
	\label{fig:system}
    \medskip
    \small
    This figure displays two Above Average and three Below Average operators. 
\end{figure}

\begin{figure}
    \centering
    \includegraphics[width=0.99\columnwidth]{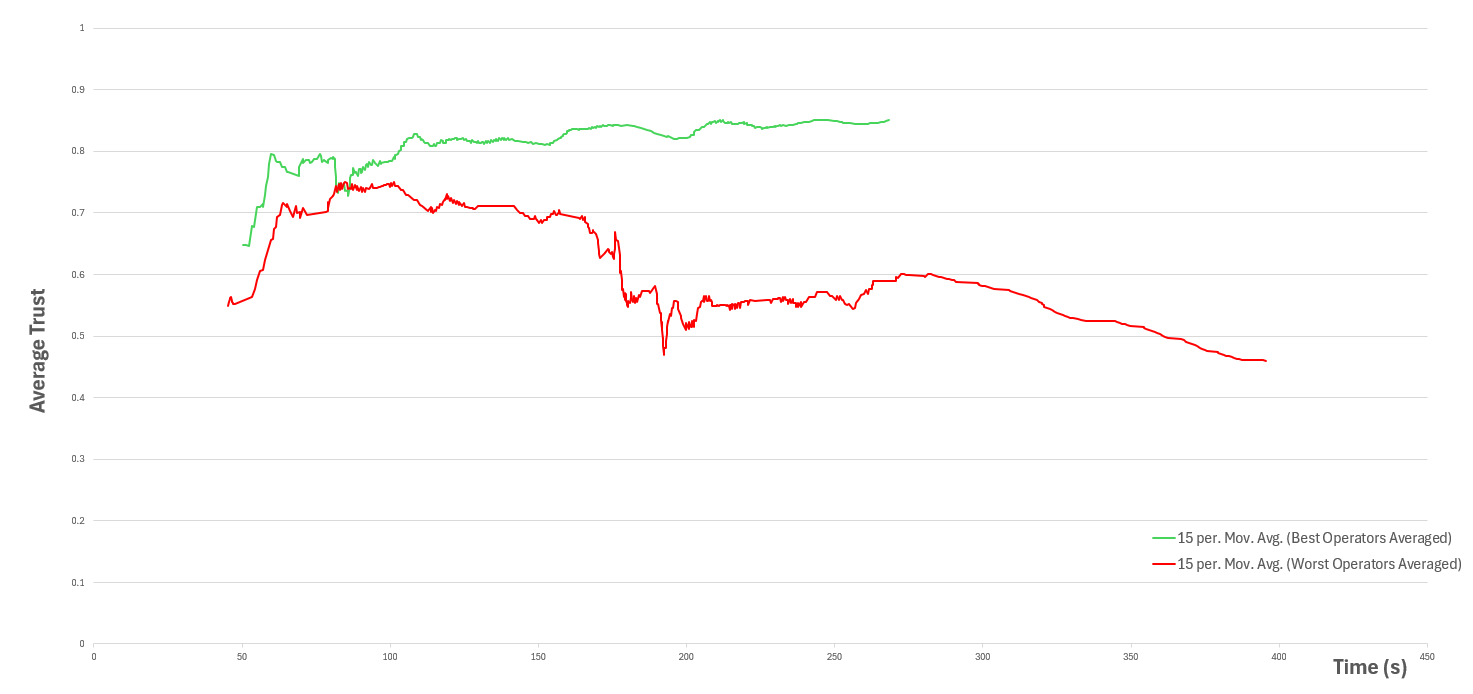}
    \caption{Comparison of the best and worst operator's estimated trust}
	\label{fig:system}
\end{figure}

Some representative cases of Above and Below-average operators can be seen in Figure 2. In Figure 3 you can see a comparison between the average Trust of the best and worst operators.

Any variance in recording trust at the beginning of the trial is due to certain operators choosing not to teleoperate the robot for that duration. This fact has been accounted for by the model, which only measured trust when the operator was teleoperating the robot.

There are cases of operators that had collisions and were placed in the middle of the strata (operator 15), or operators that completed all goals and had no collisions that were placed in a lower place in the order (operator 4) (see table in Figure 4). This signifies that while the Trust coefficient factor can affect the estimated trust towards the operator, it is also dependent on the trustworthiness or confidence metrics.

\begin{figure}
    \centering
    \includegraphics[width=0.99\columnwidth]{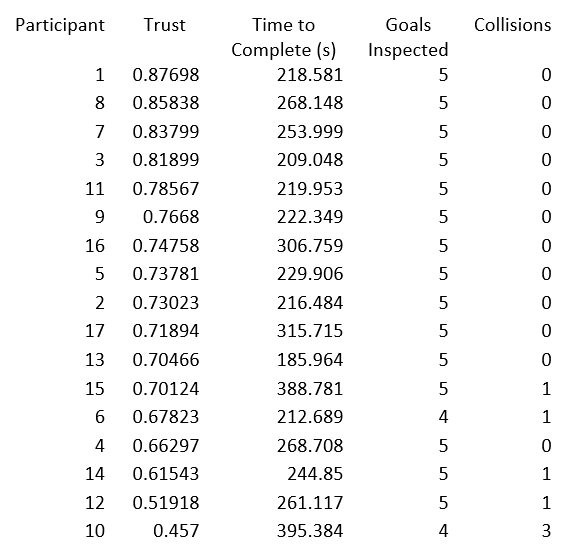}
    \caption{Operator's capability metrics sorted for Trust}
	\label{fig:system}
\end{figure}

These findings agree with our qualitative analysis. The qualitative analysis was based on notes from the experiment and the observation of the operators in each trial, lastly all the metrics collected about each operator were also taken into account.

\section{Discussion and Limitations}

The classification of the operators in the dataset by the AI agent works in a fashion similar to a human supervisor monitoring the operator in real time. Results provide evidence that the model can produce, in real time, an estimation of confidence in the operator's ability. This is something difficult for a human expert to do in real time based solely on the data collected by the system. This is partly due to the amount of parameters that have to be monitored, and the fact that humans will also rely on other factors (e.g. nonverbal cues, task-related knowledge). 


As referring to the initial research questions posed, we have reached the following answers.

\begin{itemize}

    \item A1 We implemented a wide variety of metrics with different levels of abstraction in their context. Depending on each use case and the different contexts of future implementations, the Deployers are able to choose their own metrics and their own confidence probability distribution functions, tailored to their own specific needs and the criticality of their application.
    
    \item A2 We chose a weighted average function to blend the various metrics, primarily to avoid ambiguity about the resulting trust level. Also if we went with some form of learning approach we would also need to address the issue of explainability. Future Deployers of the model can easily adjust and optimize the weights to better suit their needs. 
    
    \item A3 At this point we aim to evaluate the performance of the model and we don't currently have a similar model to compare to. Due to the nature of the study we chose to compare the resulting operator trust ranking, to the ranking produced by a supervisor evaluating the operators through both qualitative and quantitative methods.
    
\end{itemize}

This contribution can be applied in a multitude of ways. In the field of human-robot collaboration and teaming, this model can allow us to create robots that have different policies for different operators. For example, the robot could choose to disable or enable a safe teleoperation feature based on the level of trust towards the partner. At the same time, the system can monitor its partner in real-time and update its level of trust towards them. This could reduce the risks to the mission by updating its policies, or informing a supervisor that the operator is possibly fatigued. This approach could also be applied beyond robot teleoperation, for example as a way to improve the supervision of vehicle operators in industrial or commercial settings. 

This work has also helped us explore and identify limitations to our current paradigm. The biggest area of improvement is the validation of the system. The data used to test the model had all the metrics needed to allow the AI agent to evaluate the operator. However, it did not focus on collecting a large amount of data for an ad-hoc evaluation of their performance and capabilities. This made the evaluation of the metrics for the quantitative validation of the model more difficult. Having identified these issues we can create not only more robust models but also allow for a more straightforward evaluation of their performance.

The motivation for this research was to provide a working concept of our framework for AI agent Trust. The focus fell mainly on the steps and components necessary for the creation and operation of the model and less on the optimization of the specific metrics used.

The next step will be for the model to be expanded through the introduction of metrics that will provide more insight into the state, intent and actions of the operator. This data will be used to improve the performance of the model and possibly affect the interaction between the Robot and the Human. Lastly, we will introduce more complex and varied scenarios for our experimental evaluation.

\section{Conclusion}

In this paper, we presented a novel framework for the quantification of AI agent trust towards a human partner. Following principles from Theory of Mind, we used metrics that monitor the human's state, intent, and actions, in addition to their performance. By extracting context from those metrics the system can estimate a level of confidence towards the human's ability to positively contribute towards the shared goal. 

The main motivation behind this study has been to improve the interaction between humans and robots. By improving the ways the robot can perceive and understand the human we can make sure that the human will receive assistance when necessary. Through the ability to hold a mental model of others, we can decide upon the extent to which we trust them and adjust our behaviour accordingly. This way we can minimize risk and increase the rewards for our efforts by choosing trustworthy partners. 

Using this framework and focusing on HRT and robot teleoperation, we created the ATTUNE model. The ATTUNE model (Artificial Trust Towards hUmaN opErator) gave the AI agent access to confidence metrics about the human. These let it evaluate its partner's trustworthiness. The system essentially evaluates the human operator in a way similar to a human supervisor. By combining the levels of trustworthiness and observing the human's behaviour and actions, the robot was able to quantify its level of Trust towards them. 

The model was verified experimentally through a human trial dataset. The dataset contained information on operators performing an exploration and inspection task in a disaster site simulated through ROS. The ATTUNE model produced a ranking of the operators based on the perceived level of trust. The evaluation of the results was performed through both a qualitative analysis of the operators and a quantitative analysis of their performance. Through this we were able to assess the performance of our model and identify the limitations of the current paradigm, guiding our future efforts.





\section*{ACKNOWLEDGMENT}
This research was conducted in, and funded by, the Cognitive Robotics Lab, the University of
Manchester, Department of Computer Science. This work was in part supported by the US AFSOR grant AFSOR/EOARD FA9550-19-1-7002. The data used to evaluate the performance of the model were provided by the Extreme Robotics Lab, University of Birmingham, School of Metallurgy and Materials, United Kingdom.

\end{document}